\documentclass[sigconf]{acmart}

\AtBeginDocument{%
  \providecommand\BibTeX{{%
    \normalfont B\kern-0.5em{\scshape i\kern-0.25em b}\kern-0.8em\TeX}}}

 \usepackage{balance}

\usepackage[T1]{fontenc}
\usepackage{aecompl}

\copyrightyear{2021}
\acmYear{2021}
\setcopyright{acmcopyright}
\acmConference[ICMR '21] {Proceedings of the 2021 International Conference on Multimedia Retrieval}{August 21--24, 2021}{Taipei, Taiwan.}
\acmBooktitle{Proceedings of the 2021 International Conference on Multimedia Retrieval (ICMR '21), August 21--24, 2021, Taipei, Taiwan}
\acmPrice{15.00}
\acmISBN{978-1-4503-8463-6/21/08}
\acmDOI{10.1145/3460426.3463609}            
\ccsdesc[500]{Computing methodologies~Computer vision}
\ccsdesc[500]{Computing methodologies~Reconstruction}
\ccsdesc[300]{Computing methodologies~Activity recognition and understanding}

\begin{document}
\fancyhead{}

\title{Can Action be Imitated? 
Learn to Reconstruct and Transfer Human Dynamics from Videos}

\author[Y. Fu, Y. Fu, Y.-G. Jiang]{Yuqian Fu$^{1}$, Yanwei Fu$^{2}$, Yu-Gang Jiang$^{1\#}$}
\affiliation{$^1$Shanghai Key Lab of Intelligent Information Processing, School of Computer Science, Fudan University 
\country{China}}
\affiliation{$^2$School of Data Science and MOE Frontiers Center for Brain Science, Fudan University
\country{China}}
\affiliation{\{yqfu18, yanweifu, ygj\}@fudan.edu.cn
\country{}}
\thanks{$\#$ indicates corresponding author}
\renewcommand{\shortauthors}{Fu, Fu and Jiang, et al.}

\begin{abstract}
Given a video demonstration, can we imitate the action contained in this video? In this paper, we introduce a novel task, dubbed \emph{mesh-based action imitation}. The goal of this task is to enable an arbitrary target human mesh to perform the same action shown on the video demonstration. To achieve this, a novel \textbf{M}esh-based \textbf{V}ideo \textbf{A}ction \textbf{I}mitation (M-VAI) method is proposed by us. M-VAI first learns to reconstruct the meshes from the given source image frames, then the initial recovered mesh sequence is fed into mesh2mesh, a mesh sequence smooth module proposed by us, to improve the temporal consistency. Finally, we imitate the actions by transferring the pose from the constructed human body to our target identity mesh.  High-quality and detailed human body meshes can be generated by using our M-VAI. Extensive experiments demonstrate the feasibility of our task and the effectiveness of our proposed method.
\end{abstract}

\keywords{Video action imitation; 3D human reconstruction; Pose transfer}

\begin{teaserfigure}
  \includegraphics[width=0.9\textwidth]{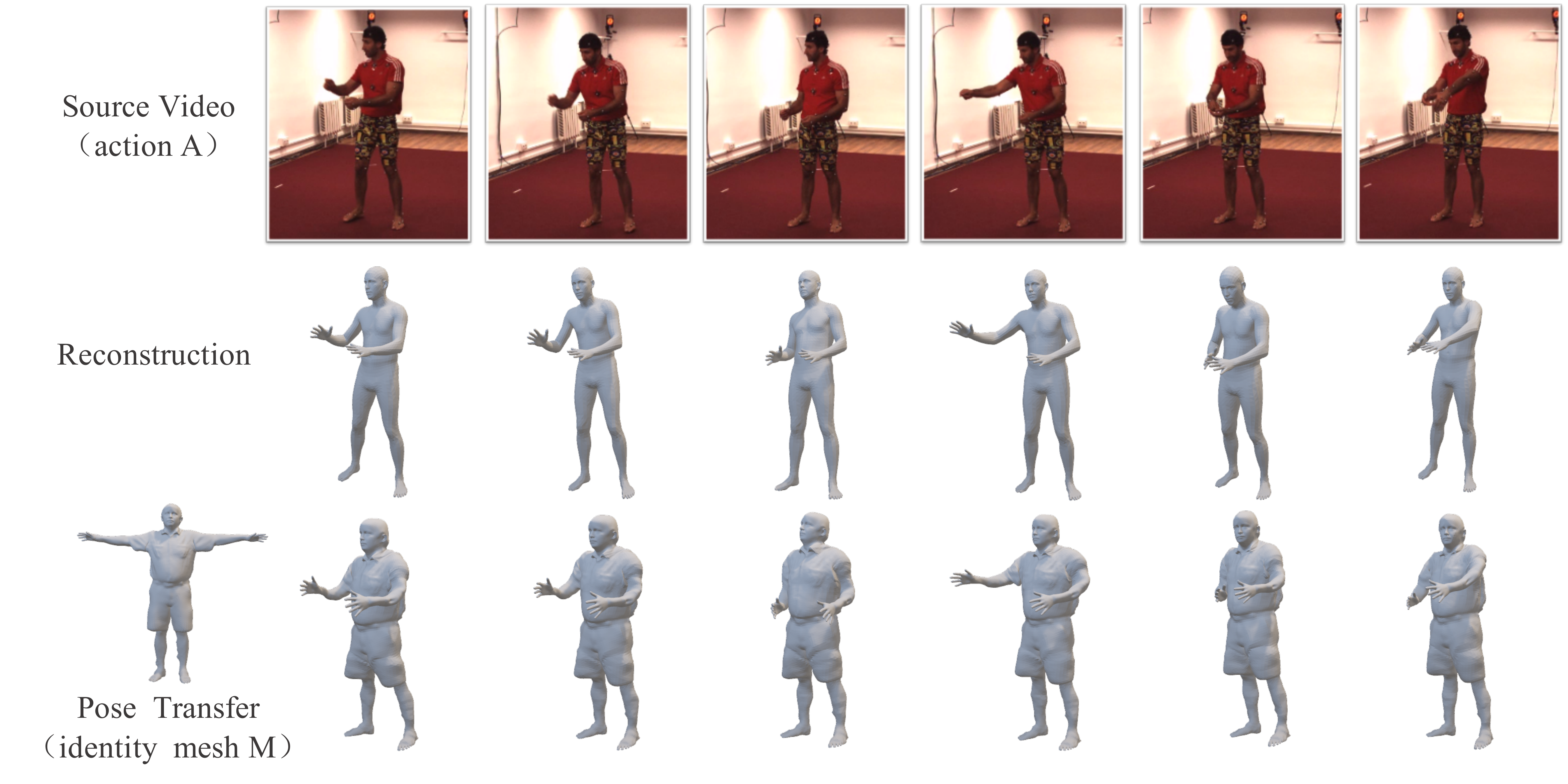}
  \caption{Examples of mesh-based action imitation. Given a source video of a person performing a certain action, we first learn to reconstruct the coherent human dynamics, and then learn to transfer the pose from the reconstructed human meshes to the target identity mesh.}
  \label{fig:teaser}
\end{teaserfigure}

\maketitle

\section{Introduction}

Synthesizing and understanding human motion plays an important role in many vision applications, such as multimedia interaction techniques, and interpretation of visual content. On the other hand, humans are very good at \textit{imitating action from video demonstrations}. The ability of imitation has been shown to be crucial for human intelligence. We believe that a 3D character should also have the ability to imitate action from a video in an intelligent system.

Despite the importance and great potential of being applied to many applications (e.g. video gaming), the task of imitating 3D action from video demonstrations remains under-explored. The most natural way to solve this task is to animate the given 3D character. However, even though lots of works have been explored in character animation \cite{macchietto2009momentum, mordatch2010robust, kavan2007skinning, wareham2008bone, bang2018spline, jacobson2011bounded,xu2020rignet,le2014robust,james2005skinning}, they can not be used to this task directly. Typically, existing character animation methods mainly research how to make the static character capable of being rigged and they are character-specific. While our task requires the methods to capture and imitate the motion dynamics in the video demonstrations automatically.

Another line of work is imitation learning \cite{hussein2017imitation,lin1992self,dixon2004learning, levine2016end,oztop2002schema,thurau2004learning,thurau2004learning} which aims at enabling an intelligent agent to imitate human behaviours. However, imitation learning mainly focuses on extracting the knowledge of how the performer acts under different environments and applying it in new surroundings. Most of the imitation learning methods are designed for robotic systems, and necessary hardware equipment or simulation environments are required.

In this paper, a novel \textbf{m}esh-based \textbf{v}ideo \textbf{a}ction \textbf{i}mitation (M-VAI) solution is proposed to bypass the difficulties aforementioned. Crucially, we propose to use the 3D mesh as the representation of the human character and introduce a new task dubbed \emph{mesh-based action imitation}. In this task, we are interested in general and daily human actions, such as greeting, jumping, and walking the dog. The ultimate goal of this task is to enable a novel human character to perform the exact same action as shown in the source videos. As shown in Figure \ref{fig:teaser}, given a source video with action $A$ and a target mesh with identity $M$, we first learn to reconstruct the human dynamics from the source video. After that, we learn to transfer the pose from the reconstructed meshes to the target identity mesh.

Our M-VAI is comprised of three steps. We first take the RGB images as the input and learn to reconstruct the 3D meshes for the initial human body contained in the video. In this work, SMPLify-X~\cite{SMPL-X:2019} and GraphCMR \cite{kolotouros2019convolutional} are employed as our reconstruction backbones to obtain the initial 3D mesh sequence.
We then propose a mesh sequence smooth module, called \emph{mesh2mesh}, which takes the initial reconstructed mesh sequence as input and aims at improving the temporal consistency of the input sequence. Due to the irregularity of the mesh, most of the mesh-based reconstruction methods adopt Graph Convolution Network~\cite{kipf2016semi} to handle the human body. In this paper, inspired by the great success of 3D convolution in video action recognition~\cite{tran2015learning, qiu2017learning}, we treat the mesh sequence as a video-like cuboid and adopt the 3D convolution to smooth the neighboring meshes. Besides, as the human action is formed by continuously changing human poses, a novel motion loss is further proposed to force the action of the smoothed mesh sequence to be consistent with the actual one. Our third step is learning to transfer the pose of the source mesh to the target identity mesh. 
Specifically, we adopt NPT~\cite{wang2020neural}, a neural pose transfer network, in our framework. Finally, for an arbitrary identity, our method is able to produce a coherent 3D mesh sequence performing the demonstrated action. Moreover, our method allows plug-and-play integration of the various human reconstruction models.

\noindent \textbf{Contributions.} 
We summarize our contributions as follows. 
1) For the first time, we introduce the task of \emph{mesh-based action imitation}.
The goal of this task is to enable an arbitrary 3D human mesh to perform the same action shown in video demonstrations automatically.  
2) We provide a solution termed M-VAI to imitate the action by learning to reconstruct and transfer human dynamics from videos. Moreover, our method is portable to various human reconstruction models.
3) We propose a mesh sequence smooth module, called \emph{mesh2mesh}, to improve the temporal consistency of the mesh sequence. Specifically, we adopt 3D convolution to model the temporal mesh sequence and introduce a \emph{motion loss} to force the mesh sequence to be consistent.
4) A large number of high-quality meshes with fine-grained geometry details can be generated 
using our model.
Extensive experiments show the feasibility of the proposed novel task and the effectiveness of our solution.

\section{Related work}

\noindent\textbf{Character Animation.} Many efforts have been done to animate characters which can be briefly divided into the there types: physics-based methods \cite{macchietto2009momentum, mordatch2010robust}, geometric-based methods \cite{kavan2007skinning, wareham2008bone, bang2018spline, jacobson2011bounded} and data-driven methods \cite {xu2020rignet,le2014robust,james2005skinning}. There are two major differences between our novel task and such animation-based methods. Firstly, these methods mainly focus on how to make the character capable of being rigged, and extra information and specific soft-wares such as Maya and Unreal Engine are required to perform the animation. While our novel task aims at teaching the character to perform the same action shown in the video demonstrations which takes advantage of the actions contained in the relatively cheap and massive video data to synthesize the desired mesh sequence. Besides, our method is platform-free. Secondly, animation-based methods are designed to rig the input character itself, it may not good at learning actions from other characters, while our method can handle this problem well.

\noindent\textbf{Imitation Learning.} Imitation learning \cite{hussein2017imitation,lin1992self,dixon2004learning, levine2016end,oztop2002schema,thurau2004learning} aims at imitating the human behaviours for robotic systems such as navigation \cite{lin1992self,dixon2004learning}, object manipulation \cite{levine2016end,oztop2002schema}, and games \cite{thurau2004learning}. Under the general paradigm of the imitation learning setting, there are some demonstrations, an agent (a learning machine), and an environment. The demonstrations are used to learn the policy which indicates how the performer acts under different surroundings. The policy is further used to guide the action of the agent in the given environment. Notably, the policy can be further refined based on the performance of the agent. Apart from learning from demonstrations, we are actually quite different. The core of imitation learning is learning the policy, and the actions generated by it are supposed to be different under various environments, while our task is more like a transfer task that transfers the pose from the performer to the target character.

\noindent\textbf{Human Pose Transfer.} Given a source data with pose and a target data with identity, the purpose of pose transfer is to transfer the pose from the source data to the target data while keeping the identity unchanged. Some pose-guided person generation methods \cite{li2019dense, neverova2018dense, ma2017pose, liu2018pose, chan2019everybody} have been explored on the 2D image domain, while 3D human pose transfer is relatively under-explored due to its complexity. NPT \cite{wang2020neural} is the recently proposed neural pose transfer network that handles 3D pose mesh and 3D identity mesh. Different from either 2D-based methods or 3D-based methods, our task takes 2D image frames which contain pose information and 3D character which represents identity as the input and aims to generate a new 3D mesh sequence. Thus, \emph{mesh} is introduced as the representation of the human body to narrow the huge gap between 2D images and 3D human bodies, and NPT is adopted as one of our key components. The most related work to us is ~\cite{guan2019human} which also transfers the action of the video demonstrations to the 3D mesh. However, since the SMPL model is used for modeling the transferred human body, the task is somewhat limited. By contrast, we aim at enabling arbitrary target mesh to imitate the actions, which is more versatile.

\noindent\textbf{3D Human Reconstruction from Images/Videos.} 3D reconstruction from images \cite{Bogo:ECCV:2016,kolotouros2019learning,kolotouros2019convolutional,hmrKanazawa17,guler2019holopose,omran2018neural,pavlakos2018learning,varol2018bodynet,SMPL-X:2019, jiang2020coherent, kulon2020weakly, choi2020pose2mesh} and videos \cite{arnab2019exploiting,huang2017towards,dabral2018learning,rayat2018exploiting,mehta2018single,pavllo20193d,mehta2017vnect, kocabas2020vibe, zanfir2020weakly} have been explored a lot. For most of the methods, SMPL model \cite{SMPL:2015}, a parametric human model, is widely used as the representation of human body. SMPLify \cite{Bogo:ECCV:2016} and SMPLify-X \cite{SMPL-X:2019} are two end-to-end methods to reconstruct human mesh by fitting the 3D keypoints of SMPL model to the 2D keypoints of image.  GraphCMR \cite{kolotouros2019convolutional} reconstructs human body by introducing a GCN \cite{kipf2016semi} to regress both the 3D locations of the vertices and the parameters of SMPL model.

We take advantage of the human body reconstruction models and use them to predict the 3D meshes from the input image frames which we insist is the most suitable way to represent the pose information. Since video-based methods e.g. \cite{kocabas2020vibe, zanfir2020weakly} assume massive in-the-wild human motions in SMPL format are available which may not be realistic in the real world, thus we stick to the image-based methods. Technically, the plug-and-play mesh2mesh is proposed to improve the temporal consistency of the mesh sequence without additional meshes as training data. What worth mentioning is that it is the first time the concept of "mesh cuboid" is proposed which treats the complex and geometric mesh as a normal cuboid. This may bring something new to the representation of 3D mesh. 


\section{Methodology}

\renewcommand{\thefigure}{2}
\begin{figure*}[h!]
\centering
	\includegraphics[width=0.8\linewidth]{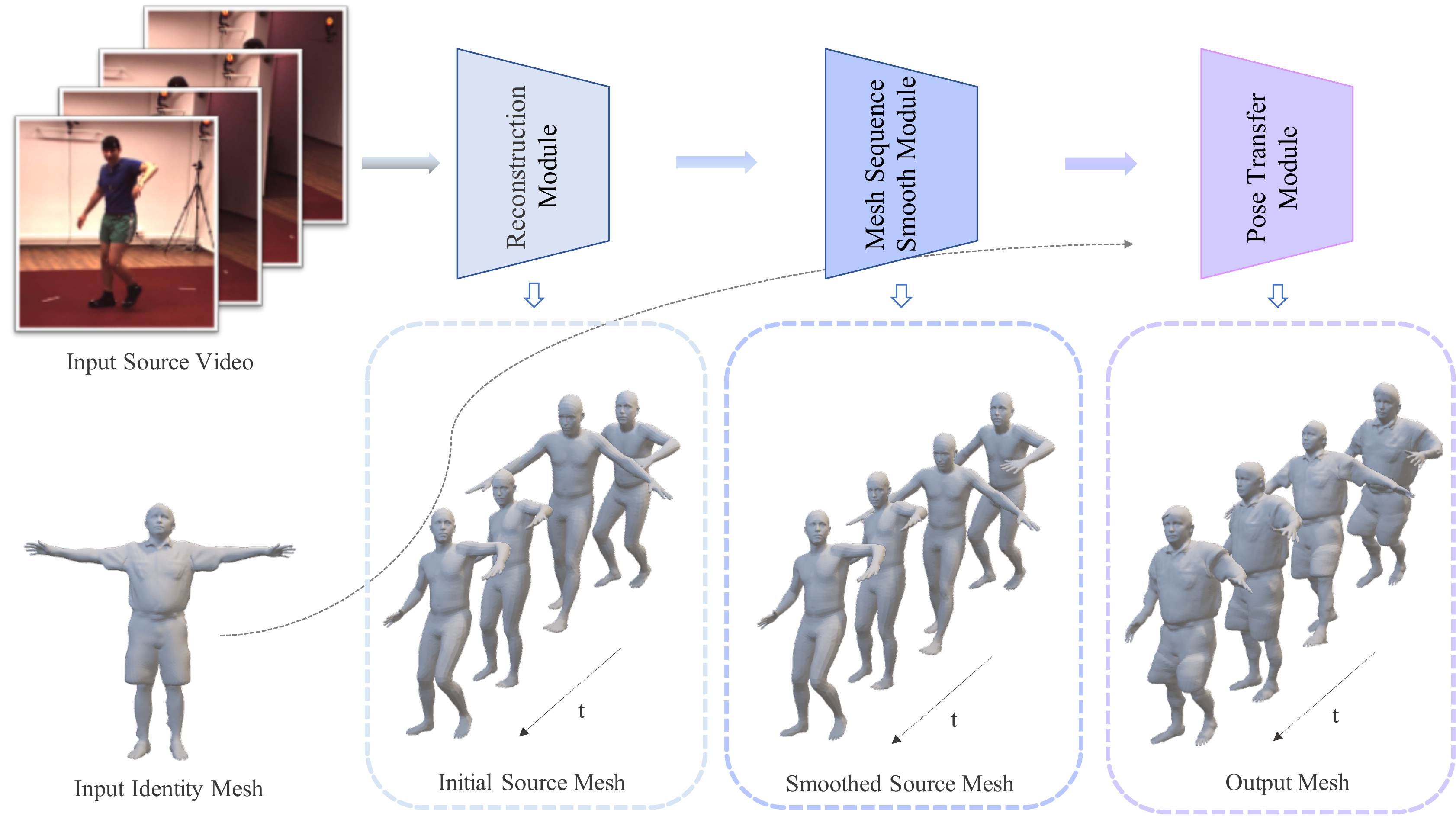} 
	\caption{Overview of our method. Given an input source video, a reconstruction module is first used to predict the initial source meshes. After that, the initially reconstructed meshes are fed into the mesh sequence smooth module which aims at improving the temporal consistency of the mesh sequence. Finally, for each mesh in the smoothed source mesh sequence, a pose transfer module is utilized to transfer the pose from it to the input identity mesh. In this way, we generate the final output mesh sequence which performs the same action as that of the input source video while keeps the identity maintained.}
	\label{fig:pipeline} 
\end{figure*}

The overall framework of our mesh-based video action imitation method (M-VAI) is summarized in Figure \ref{fig:pipeline}. M-VAI is mainly composed of three components: reconstruction module $\mathcal{R}$, mesh sequence smooth module $\mathcal{F}$, and pose transfer module $\mathcal{G}$. Given an input source video $V = \{I_t\}_{t=1}^{T} $ which contains $T$ frames, we obtain the initial source mesh $ \tilde{Y} = \{\tilde{M}_t\}_{t=1}^{T}$ using the reconstruction module $\mathcal{R}$ which takes image sequence as input and outputs the corresponding mesh sequence. Then, these initially reconstructed meshes are fed into the mesh sequence smooth module $\mathcal{F}$ proposed by us to make the meshes more coherent. So far, we reconstruct the smoothed source meshes $ \hat{Y} = \{{\hat{M}}_t\}_{t=1}^{T}$  from the input video, and the dynamics remain consistent with the original video. On this basis, for each mesh $\hat{M}_t$ in the smoothed mesh sequence, we transfer its pose to that of the input identity mesh $M_{id}$ by employing the pose transfer network $\mathcal{G}$ which takes the pair of $<\hat{M}_t$, $M_{id}>$ as input and generates the mesh ${M^*}_t$ inheriting the pose from source mesh and the identity from the identity mesh. We insist that action is formed by a continuously changing human pose, thus the final output mesh sequence $Y^* = \{{M^*}_t\}_{t=1}^{T}$ is considered as imitating the action of the initial source video while maintaining the identity of the target identity human body.

\noindent \textbf{Human body representation: }  
Benefiting from the rapid development of image-based human reconstruction, the 3D mesh is introduced as the representation of the human body.  Specifically, we encode a 3D mesh $M$ using the SMPL \cite{SMPL:2015} model.  SMPL model is parameterized by a set of shape parameters $\alpha$  and a set of pose parameters $\beta$. SMPL-based 3D mesh contains 6890 vertices, formally, $M \in {\mathcal{R}^{6890\times 3}}$. More details can be found in SMPL \cite{SMPL:2015}.

Next, in section \ref{sec:recons} we briefly describe the reconstruction module. In section \ref{sec:mesh_seq} we focus on the mesh2mesh, the 3D convolution-based mesh sequence smooth module proposed by us, which is responsible to deform the vertex coordinates of meshes towards a more coherent sequence. Then, the pose transfer module is briefly described in section \ref{sec:npt}.

\subsection{Reconstruction module\label{sec:recons}}
Since the development of human body reconstruction, many flagship models have been proposed in the 3D vision community. Among them, SMPLify-X \cite{SMPL-X:2019}  and GraphCMR \cite{kolotouros2019convolutional} are typical methods proposed recently and have good quantitative and qualitative results.

SMPLify \cite{Bogo:ECCV:2016} is an end-to-end method that reconstructs 3D human mesh by fitting the 3D keypoints of SMPL to the 2D keypoints predicted from the input image. SMPLify-X\cite{SMPL-X:2019} is an improved method of SMPLify\cite{Bogo:ECCV:2016}, which proposes a new neural network pose prior encoder and a new interpenetration penalty. GraphCMR \cite{kolotouros2019convolutional} mainly constructs a Graph CNN to encode the mesh structure so that the topology of the mesh is maintained. After that, under the supervision of the ground truth vertex locations, the coordinates of the vertices can be regressed by performing graph convolution. In this paper, both SMPLify-X \cite{SMPL-X:2019} and GraphCMR \cite{kolotouros2019convolutional} are introduced as our reconstruction modules.

\subsection{Mesh sequence smooth module}
\label{sec:mesh_seq}

The initial meshes reconstructed by the reconstruction module $\mathcal{R} $ may look good in a single frame, but there will still be some undesired phenomenons. For example, a relatively poor mesh may appear in several normal-shaped meshes, or the legs of the same person may have different lengths. To address these problems, a novel mesh sequence smooth module $\mathcal{F}$, namely \emph{mesh2mesh}, is proposed by us.

Our module takes the mesh sequence as input and aims at generating a mesh sequence whose action dynamics are as consistent as possible with that of the original video. To better fuse the temporal information of the mesh sequence, we treat the sequence of mesh as a video-like cuboid and introduce the 3D convolution to deform the vertex coordinates toward a better smooth mesh sequence. 


\noindent\textbf{3D convolution:} 3D convolution has been proved to be very useful in extracting video spatio-temporal representation~\cite{tran2015learning, qiu2017learning}. As shown in Figure \ref{fig:mesh_model}(a), given a video cuboid $V \in {\mathcal{R}^{T \times H \times W}}$, the 3D kernel ${d_T}\times{d_H}\times{d_W}, ({d_T} < T) $ will slide among the spatial and temporal dimensions at the same time. Hence, it is well-suited to model the temporal information. Besides, applying 3D convolution on a cuboid results in another cuboid, preserving the volume of the input cuboid unchanged. These two characteristics together make the 3D convolution a good choice to deform the sequence data.

\noindent\textbf{Mesh cuboid:} Inspired by the video representation learning, the concept of mesh cuboid is first proposed by us. As shown in Figure \ref{fig:mesh_model}(b), given a mesh sequence, for each mesh $\tilde{M}_{t}$, we first extract its vertices and then flatten it into an N $\times$ 3 matrix. In this way, the input mesh sequence is represented as a mesh cuboid $ \tilde{Y} = \{\tilde{M}_t\}_{t=1}^{T} \in {\mathcal{R}^{T\times N\times 3}}$. Compared with the traditional video cuboid $V\in{\mathcal{R}^{T\times H \times W}}$, the number of vertices $N$ and the dimension of vertex coordinates $3$ are equivalent to the length $H$ and the width $W$, respectively. The number of meshes $T$ is naturally considered to be the depth of the cuboid, which usually refers to the number of frames as for video. The channel of each "pixel" in the cuboid is set as 1.

\noindent \textbf{Mesh2mesh module:} As shown in Figure \ref{fig:mesh_model}(c), our mesh sequence smooth module is mainly composed of eight stacked 3D convolution layers. To make sure the volume of the output mesh cuboid is the same as the input mesh cuboid, the size of all kernels is set as $5\times 1\times 3$, the stride and padding are set as $(1,1,1)$ and $(2,0,1)$ respectively. And the unsqueeze and squeeze layers are added to handle the channel of each "pixel" in the cuboid.

\noindent \textbf{Loss function:} During the training phrase, let $ J = \{X_t\}_{t=1}^T \in {\mathcal{R}^{T\times k\times 3}}$ denotes the 3D keypoints of the ground truth human bodies, where $k$ means the number of 3D joints.  Our predicted 3D joints $\hat{J} = \{\hat{X}_t\}_{t=1}^T \in {\mathcal{R}^{T\times k\times 3}} $ can be regressed from the predicted 3D meshes $\hat{Y}$ by employing the regressor provided by the SMPL \cite{SMPL:2015} model. Then, we compute the keypoints loss $\mathcal{L}_{j3d}$ as follows:
\begin{equation}
\mathcal{L}_{j3d}=\frac{1}{T} \times \sum_{t=1}^{T} \sum_{i=1}^{k} \|\hat{X}_{t,i}-X_{t,i}\|.
\label{equa:1}
\end{equation}

Besides, a novel motion loss is proposed to force the action to be consistent with the ground truth. Specifically, we insist that the movement of adjacent frames between the predicted mesh sequence and the real one should be the same. Thus, ours motion loss $\mathcal{L}_{motion} $ is defined as: 
\begin{equation}
\mathcal{L}_{motion}=\frac{1}{T} \times \sum_{t=2}^{T}\sum_{i=1}^{k}  \| \left( \hat{X}_{t,i}-\hat{X}_{t-1,i} \right) - \left( X_{t,i}-X_{t-1,i} \right) \|, 
\label{equa:2}
\end{equation}
\noindent where $\| \cdot \|$ means a per-keypoint L2 loss.  
Our complete loss $\mathcal{L} $ of $\mathcal{F}$ is:
\begin{equation}
\mathcal{L} = \mathcal{L}_{j3d} + \mathcal{L}_{motion}
\label{equa:2}
\end{equation}

\renewcommand{\thefigure}{3}
\begin{figure*}[h!]
\centering
		\includegraphics[width=0.8\linewidth]{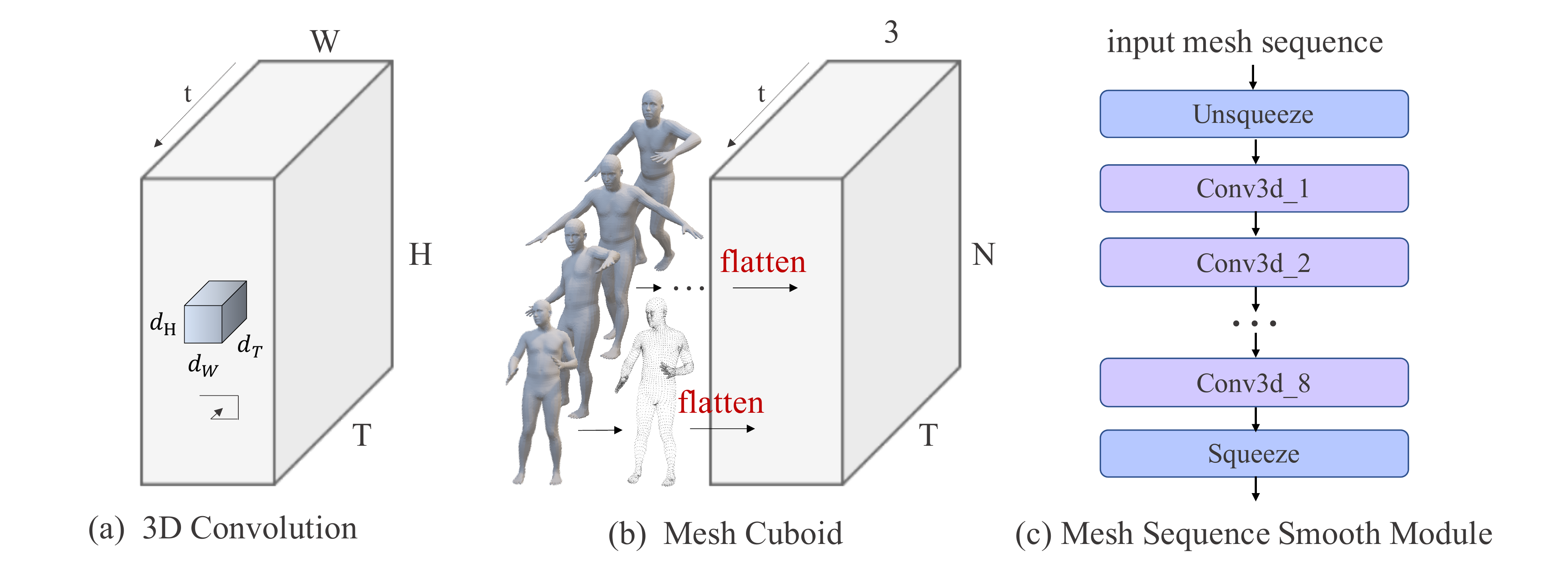} 
	\caption{Illustration of our mesh sequence smooth network. (a) Given a cuboid $V \in^{\mathcal{T\times H \times W}}$, the 3D convolution is operated by sliding the 3D kernel ${d_H}\times{d_W}\times{d_T}, ({d_T} < T)$ over the temporal and spatial dimensions. (b) We treat the sequence of mesh as a mesh cuboid $\tilde{Y} \in {\mathcal{R}^{T\times N \times 3}}$ to model the temporal information. $T$, $N$, and $3$ denotes the number of meshes, the number of vertices contained in a single mesh, and the dimension of vertex coordinates, respectively. (c) Our mesh sequence smooth model is mainly composed of eight 3D convolution layers. The unsqueeze and squeeze layers are added to make sure the volume of the cuboid is reasonable.} 
	\label{fig:mesh_model} 
\end{figure*}

We highlight some advantages of our \emph{mesh2mesh} module: 
1) It is designed as a plug-and-play mesh sequence smooth module. In other words, it can be easily applied to existing reconstruction modules. 
2) The mesh cuboid only considers optimizing the locations of the vertices, and represents them in the regular Euclidean space, making it possible to handle the complex mesh in an easy way. 
3) We take advantage of 3D convolution in extracting spatial-temporal information and use it to smooth the mesh sequence. Most importantly, we do not change the volume size of the input data which makes it portable. 
4)  To guide the action dynamics, the motion loss is first proposed by us. 
It guides the movement of the predicted mesh sequence to be the same as the ground truth.

\subsection{Pose transfer module}
After generating the coherent mesh sequence through the mesh2mesh module, we propose the pose transfer module by employing the NPT~\cite{wang2020neural} to transfer the pose of the source mesh to the target identity mesh.

We adopt \emph{max pooling} to produce a latent vector to represent the pose feature and remove the \emph{instance normalization} layer in the first \emph{SPAdaIN ResBlock}, and here we call this pose transfer module $\mathcal{G}$. Given 
the identity mesh $M(\alpha_{id},\beta_{id})$ and pose meshes $\{{\hat{M}}_t(\alpha_{pose},\beta_{t})\}_{t=1}^{T}$, we can get 
\begin{equation}
\{M^*_t(\alpha_{id},\beta_{t})\}_{t=1}^{T}=\mathcal{G}(\{{\hat{M}}_t(\alpha_{pose},\beta_{t})\}_{t=1}^{T},M(\alpha_{id},\beta_{id}))
\label{equa:npt}
\end{equation}
\noindent where $\alpha$ represents the human shape and $\beta$ represents the human pose. 
Here $\alpha$ is beyond SMPL's~\cite{SMPL:2015} shape parameters.

By using \emph{max pooling}, we can get the encoded representation of the `action' from one clip of the video. 
We argue that getting this action vector is a more general method in our task because we hope to handle different kinds of 3D meshes and the number of points may be different. In this way, we can copy this vector for arbitrary times and match the number of points of the identity mesh, breaking the limitation of the original NPT~\cite{wang2020neural} method which needs the number of vertices to be the same.
Thus, we are able to handle the complex identity meshes even with clothes on. To the end, by taking the relatively cheap videos which contain all kinds of actions, we can imitate the actions from them to our target identity meshes and generate massive fine-grained 3D meshes sequence with the desire dynamics. 

\label{sec:npt}



\renewcommand{\thefigure}{4}
	\begin{figure*}
	\begin{centering}
		\includegraphics[width=0.8\linewidth]{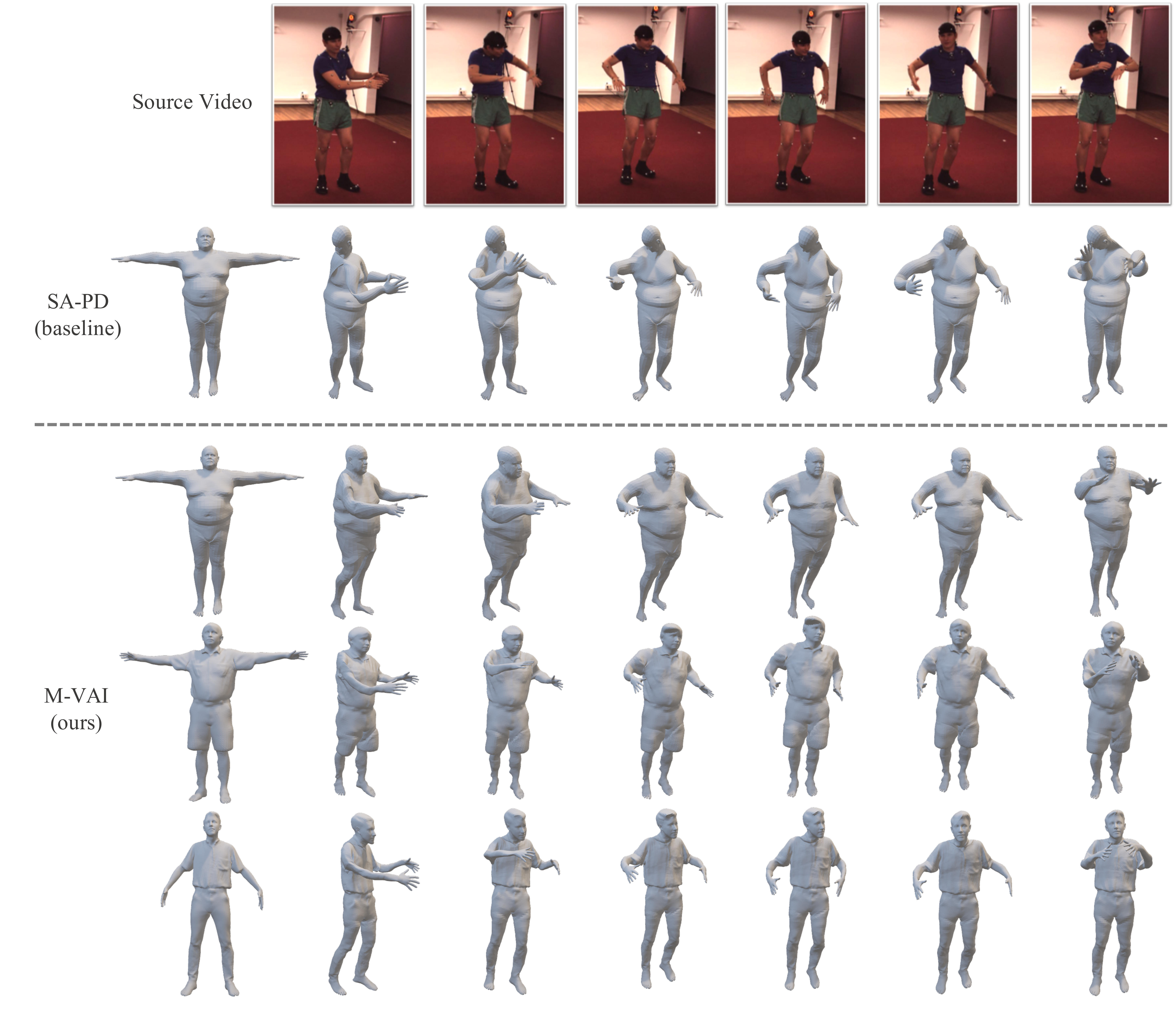} 
		\par\end{centering}
	\caption{Results of action imitation. The first row is the source video that provides the action, the second row shows the deformation results of SA-PD baseline and the last three rows are the results of our M-VAI with different identity meshes as input.}
	\label{fig:main_result_3dv} 
\end{figure*}

\renewcommand{\thefigure}{5}
\begin{figure}[h!]
\centering
		\includegraphics[width=0.85\linewidth]{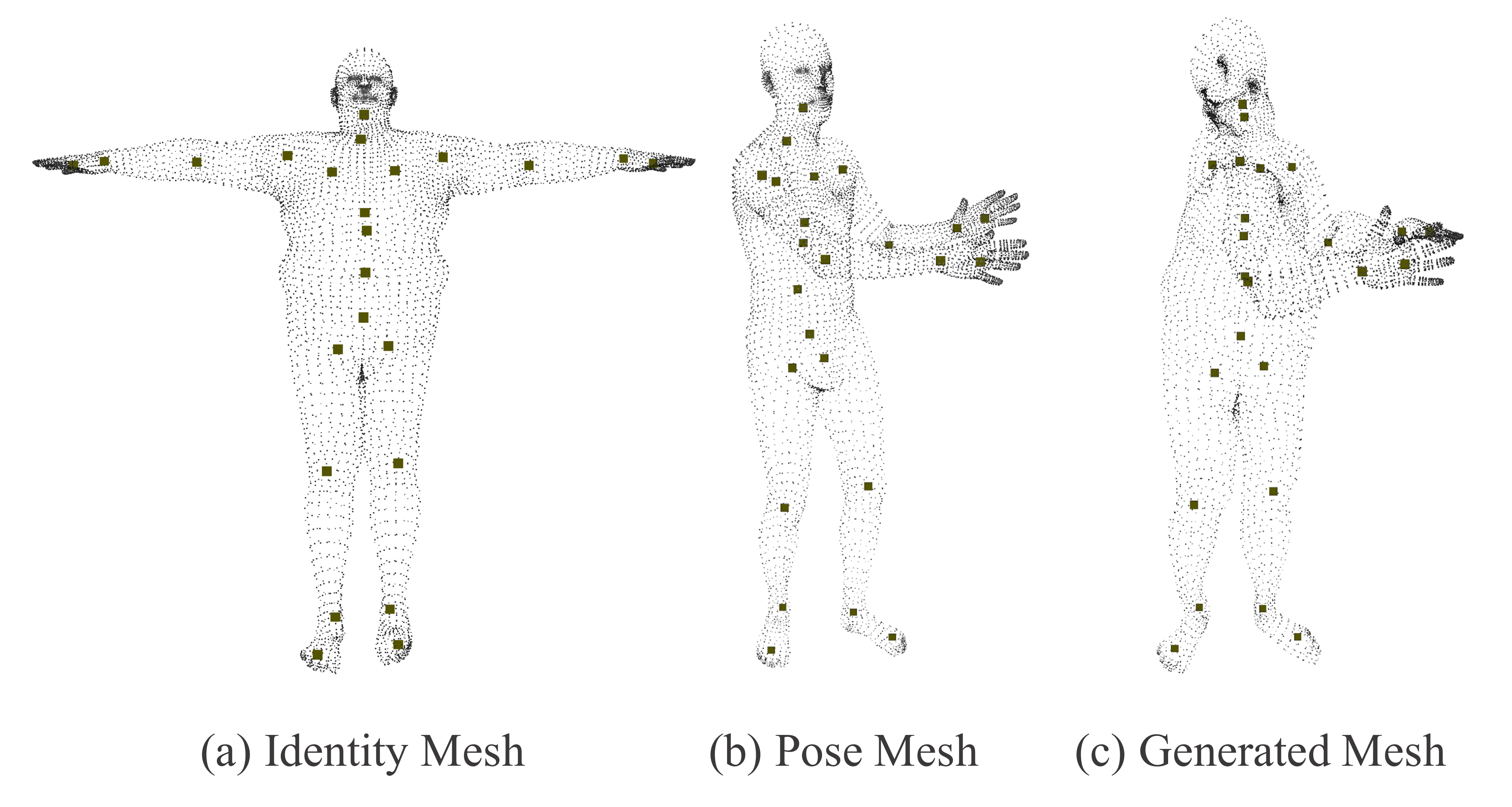} 
	\caption{An example of SA-PD is shown. The smaller dots represent the vertices of the mesh, while the bigger dots represent the 3D joints of the mesh. }
	\label{fig:skel} 
\end{figure}

\section{Experiment}

In this section, we first introduce the experimental setup including datasets, video processing, evaluation metrics, and implementation details in sec \ref{sec:setup}. Then, we mainly demonstrate the results of action imitation, our mesh2mesh module, and ablation study in sec \ref{sec:action_imitation}, \ref{sec:mesh2mesh}, and \ref{sec:abla}, respectively.

\subsection{Experimental setup\label{sec:setup}}
\noindent\textbf{Datasets.}
We conduct experiments on Human3.6M \cite{ionescu2013human3} which provides the raw video and the corresponding 3D keypoints ground truth. Human3.6M~\cite{ionescu2013human3} is an indoor dataset, recording actors performing different actions such as greeting, walking, and waiting. We use the subjects S1, S6, S7, S8 as the training set, S9 and S11 as the test set. 

\noindent\textbf{Video processing.} We down-sample the video with a sampling frequency of $1/25$. During the training phase, we randomly sample a continuous clip of $T$ frames. During the testing phase, the middle $T$-frame clip is selected as the input data.

\noindent\textbf{Reconstruction modules.} In order to validate the versatility of the method, both SMPLify-X~\cite{SMPL-X:2019} and GraphCMR~\cite{kolotouros2019convolutional} described in \ref{sec:recons} are employed as our reconstruction modules. We first get the 2D keypoints of images by running OpenPose~\cite{8765346}, then we obtain the initial mesh sequence through SMPLify-X and GraphCMR. Since they both have been trained on the training set of Human3.6M,  we directly use the pre-trained model provided by SMPLify-X \cite{SMPL-X:2019,smplifyxcode} and GraphCMR \cite{kolotouros2019convolutional, cmrcode}, respectively.

\noindent\textbf{Implementation details.} As for the mesh2mesh and pose transfer modules, we employ a two-step training procedure. We first train the mesh2mesh module $\mathcal{F}$ to obtain a more consistent mesh sequence. We then train the neural pose transfer module $\mathcal{G}$ to transfer the dynamics from the source human body to the target identity one. Specifically, Adam optimizer \cite{kingma2014adam} with a learning rate of $3\times 10^{-3}$ is used for $\mathcal{F}$. We train the Pose Transfer Module by creating 3D meshes of 16 identities with 800 poses created by SMPL~\cite{SMPL:2015} with the same setting used in NPT~\cite{wang2020neural}. The $T$ is set as 16.

\renewcommand{\thefigure}{6}
\begin{figure*}[h]
	\begin{centering}
		\includegraphics[width=0.85\linewidth]{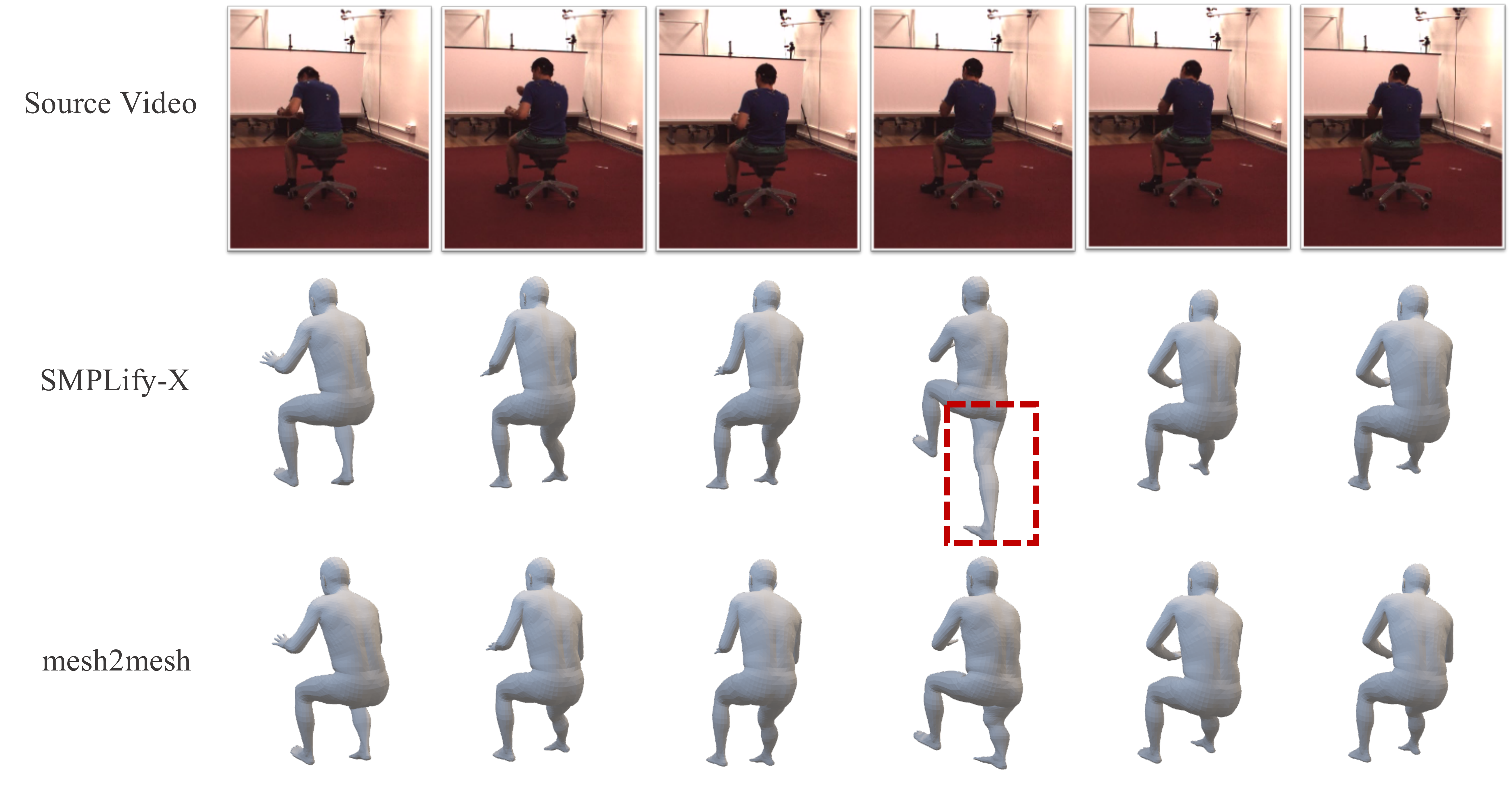} \vspace{-0.1in}
		\par\end{centering}
	\caption{Results of mesh2mesh. 
	The first row is the source video. The second row shows the reconstruction results of SMPLify-X, and the last row shows our results of mesh2mesh. Our mesh2mesh sequence smooth network can generate more continuous and correct meshes.}
	\vspace{-0.1in}
	\label{fig:mesh2mesh-result} 
\end{figure*}

\noindent\textbf{Evaluation metrics.} The novel mesh-based action imitation proposed by us belongs to a generation task, and there is no available ground truth, thus we can only evaluate it qualitatively. As for the human body reconstruction, the mean per joint position error (MPJPE) defined in \cite{zhou2018monocap} is used to measure the performance of reconstruction modules. In this paper, the average MPJPE of $T$ frames contained in the same clip is reported.

\subsection{Results of action imitation\label{sec:action_imitation}}

\noindent\textbf{Animation-based baseline.} We employ a \textbf{s}keleton-based \textbf{p}ose \textbf{d}eformation method as our \textbf{a}nimation-based baseline, which can be called as \emph{SA-PD}. First, we reconstruct the 3D mesh sequence for the input image frames to get the 3D pose information by using SMPLify-X as our M-VAI does. Then, for both identity mesh and pose meshes, the 3D joints of the human body are obtained by applying the regressor provided by the SMPL human model. The regressed joints are further used to recover the skeleton of human bodies with prior knowledge of the SMPL model. Now, both the skeleton of identity and pose meshes have been known. 

Given a pose mesh and an identity mesh, we take 3 steps to deform the identity mesh towards the pose mesh. 1) Skeleton Alignment: we denote the skeleton of pose and identity as $skel_{pose}$ and $skel_{id}$, respectively. We first align the root joint of $skel_{pose}$ to that of $skel_{id}$, then for bones of $skel_{pose}$ whose parent node is the root, we calculate the normalized direction vectors which indicating the local pose for them. The calculated direction vector and the length of the corresponding bone of $skel_{id}$ are used to infer the new bone of $skel_{pose}$. Repeat this progress until all the bones in the original $skel_{pose}$ have been aligned. In this way, we try to reduce the negative effects caused by skeleton mismatch, including global location offset and different lengths of the limb. 2) Calculating Transformation Matrix: since the skeleton structure of pose mesh and identity mesh have been obtained, we calculate the transformation matrix between $skel_{pose}$ and $skel_{id}$ in the local coordinate system. 3) Skinning Deformation: we calculate the binding weights of Liner Blend Skinning (LBS) \cite{jacobson2011bounded} for identity mesh by using the tools provided by \cite{pinocode}. Finally, we deform the identity mesh according to the binding weights and the transformation matrix by LBS \cite{jacobson2011bounded}.

\noindent\textbf{Qualitative results.} The results of animation-based baseline SA-PD and our M-VAI are shown in Figure \ref{fig:main_result_3dv}. The first row shows the image frames of the source video selected from the test set of the Human3.6M dataset. The final mesh sequences generated by SA-PD and M-VAI are visualized in the following rows. Among these, the identity meshes of the second row and third row are the same which is an SMPL-based human body. The identity meshes wearing clothes of the last two rows are selected from MG-dataset \cite{bhatnagar2019multi} which is much more challenging.

We highlight several important results: 1) We observe that SA-PD can imitate the approximate pose, but fails to recover the details such as head and hands. In order to better clarify this phenomenon, we visualize a pair of $\left\langle identity \ mesh, pose \ mesh, generated \ mesh \right\rangle$ in Figure \ref{fig:skel}. The smaller dots represent the vertices of the mesh, while the bigger dots denote the 3D joints. We observe that the skeleton (joints) in the generated mesh is factually close to that pose mesh, however not all the vertices can be deformed well especially when they are far away from the joints. 2) Our M-VAI successfully transfers the dynamics from the input image frames to our identity mesh. It turns out that our mesh-based imitation method outperforms the skeleton-based animation method by a large margin which can be observed by comparing the result of the second row and the third row. 3) Since the skeleton-based animation method requires the skeleton of the pose mesh and the identity mesh to be the same, thus it can not handle the more complex meshes selected from MG-dataset \cite{bhatnagar2019multi} which have more than 20,000 vertices. While our M-VAI is not limited and generates mesh sequence well which can be observed in the last two rows. What worth mentioning is that it makes us possible to synthesize massive meshes with fine-grained details.

\subsection{Results of mesh sequence smooth module \label{sec:mesh2mesh}}

\begin{table}
\begin{centering}
\setlength{\tabcolsep}{0.24em}
\renewcommand{\arraystretch}{1.2}
\small
\begin{tabular}{c| c | c }
\hline 
  & SMPLify-X\cite{SMPL-X:2019} & GraphCMR\cite{kolotouros2019convolutional} \tabularnewline
\hline  
backbone &  136.4 &  74.7 \tabularnewline
backbone + mesh2mesh &  128.4 $(8.0 \downarrow)$ &  \textbf{72.8} $(1.9 \downarrow)$ \tabularnewline

\hline 
\end{tabular}
\par\end{centering}
\caption{Evaluation of human body reconstruction in Human3.6M~\cite{ionescu2013human3}. 
The MPJPE errors in mm are reported. 
Both the performance of SMPLify-X~\cite{SMPL-X:2019} and GraphCMR~\cite{kolotouros2019convolutional} are improved by applying mesh2mesh.}
\label{tab:error_tab} 
\vspace{-0.1in}
\end{table}

\renewcommand{\thefigure}{7}
\begin{figure}
	\begin{centering}
		\includegraphics[width=0.85\linewidth]{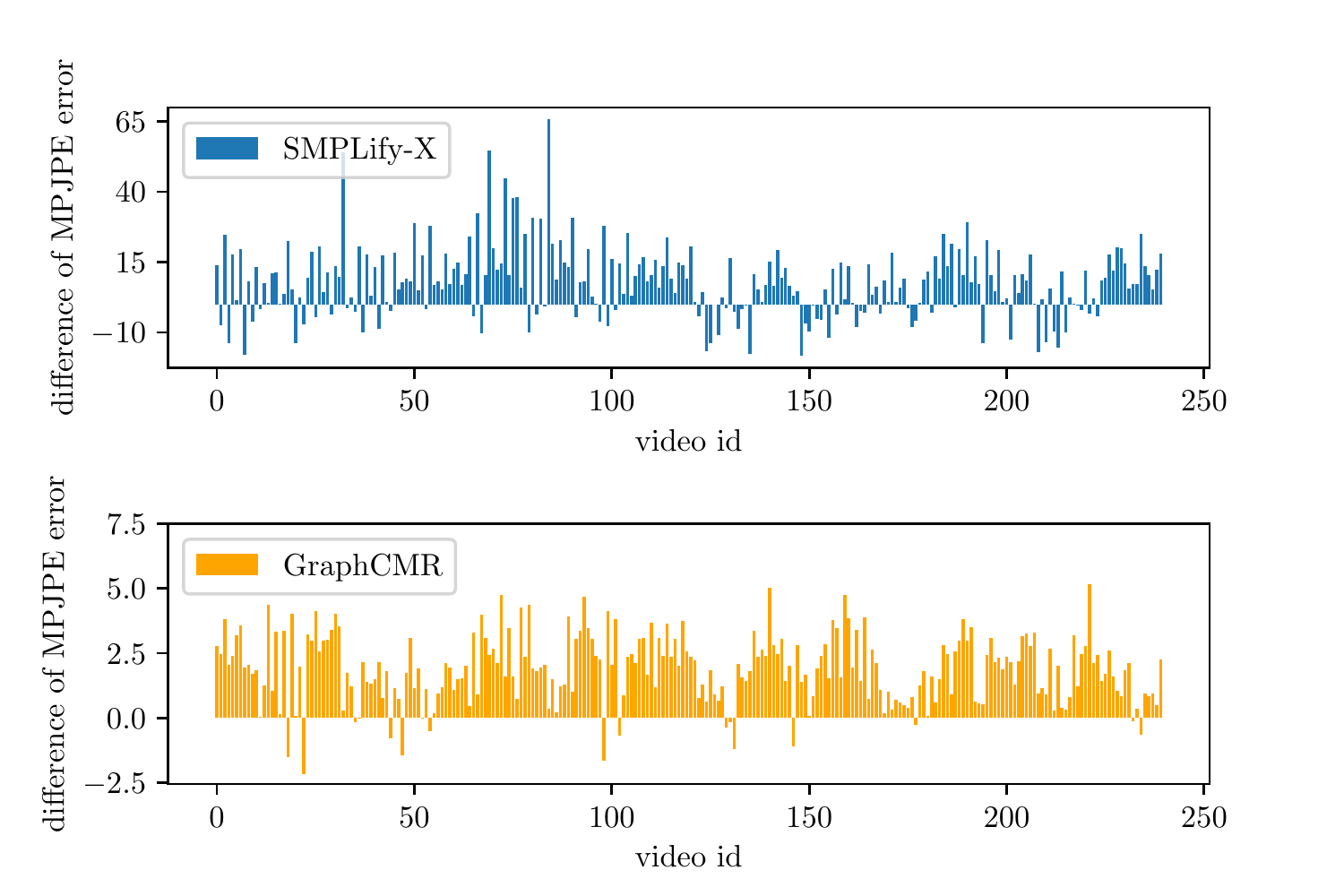} 
		\par\end{centering}
	\caption{MPJPE error for each video in the test set. 
		Our mesh2mesh (orange) outperforms the backbones (blue) in most cases. }
	\label{fig:error_fig} 
\end{figure}

We conduct experiments based on SMPLify-X~\cite{SMPL-X:2019} and GraphCMR~\cite{kolotouros2019convolutional} and report the MPJPE errors in Table \ref{tab:error_tab}. The results of the backbones are evaluated by us. 

We can see that after applying the mesh2mesh module, the performance of the backbone modules is improved. For SMPLify-X~\cite{SMPL-X:2019}, the MPJPE error is reduced from 136.4 to 128.4, while for GraphCMR~\cite{kolotouros2019convolutional} the MPJPE error is reduced from 74.7 to 72.8.  

In addition, we also analyze the impact of mesh2mesh on each video contained in the test set as shown in Figure \ref{fig:error_fig}. The horizontal axis represents different videos, and the vertical axis represents the MPJPE error difference between the original backbone and the backbone with the mesh2mesh module. A positive value indicates that our mesh2mesh module improves the performance of the backbone. Whether for SMPLify-X~\cite{SMPL-X:2019} or GraphCMR~\cite{kolotouros2019convolutional}, a performance gain is brought by our mesh2mesh in most cases.

To have a more intuitive feeling, we visualize a typical example in Figure \ref{fig:mesh2mesh-result}. We demonstrate the source video in the first row, the mesh reconstructed by SMPLify-X~\cite{SMPL-X:2019} and mesh2mesh in the second and third rows, respectively. Even the neighboring meshes are correctly reconstructed, SMPLify-X~\cite{SMPL-X:2019} fails to reconstruct the correct pose for the fourth frame. This inconsistency is improved by our mesh2mesh, which indicates that our mesh2mesh does make use of the temporal information and then smooth the mesh sequence towards a more consistent sequence.

\subsection{Ablation Study} \label{sec:abla}
To validate the effectiveness of our components, we conduct some ablation studies on the motion loss and the mesh2mesh module. Generally, we evaluate our methods on the Human3.6M \cite{ionescu2013human3} dataset, and the MPJPE errors defined in \cite{zhou2018monocap} are reported. Unless special instructions are given, all the training details are exactly the same as we described in the main text.

\noindent\textbf{Ablation study of motion loss.} As shown in Table \ref{tab:motion-loss}, the results of with/without motion loss are reported. We observe that for both SMPLify-X \cite{SMPL-X:2019} and GraphCMR \cite{kolotouros2019convolutional}, the performance is improved after employing the motion loss. Specifically, the MPJPE error decreases from 134.9 to 128.4 on SMPLify-X, from 73.2 to 72.8 on GraphCMR, which indicates that the motion loss proposed by us does force the generated mesh sequence to be more consistent with the ground truth.

\begin{table}[h]
\begin{centering}
\setlength{\tabcolsep}{0.24em}
\renewcommand{\arraystretch}{1.2}
\small
\begin{tabular}{c| c | c }
\hline 
  & SMPLify-X\cite{SMPL-X:2019} & GraphCMR\cite{kolotouros2019convolutional} \tabularnewline
\hline  
without motion loss &  134.9 &  73.2 \tabularnewline
with motion loss (ours) &  \textbf{128.4} $(6.5 \downarrow)$ &  \textbf{72.8} $(0.4 \downarrow)$ \tabularnewline
\hline 
\end{tabular}
\par\end{centering}
\caption{Ablation study of motion loss. The MPJPE errors in mm are reported. After employing the motion loss, the performance is improved. }
\label{tab:motion-loss} 
\vspace{-0.1in}
\end{table}

\begin{table}[h]
\begin{centering}
\setlength{\tabcolsep}{0.24em}
\renewcommand{\arraystretch}{1.2}
\small
\begin{tabular}{c| c }
\hline 
kernel size & \ \ \ SMPLify-X\cite{SMPL-X:2019} \ \ \ \tabularnewline
\hline  
$5 \times 3 \times 3$ &  179.0  \tabularnewline
$3 \times 1 \times 3$ &  131.3  \tabularnewline
$5 \times 1 \times 1$ &  134.6  \tabularnewline
$5 \times 1 \times 3$ (ours) &  \textbf{128.4}  \tabularnewline
\hline 
 number of convolutional layers \ & \ \ \ SMPLify-X\cite{SMPL-X:2019} \ \ \  \tabularnewline
\hline  
3 layers &  129.5  \tabularnewline
8 layers (ours) &  \textbf{128.4} \tabularnewline
12 layers &  138.8 \tabularnewline
\hline
\end{tabular}
\par\end{centering}
\caption{Ablation study of mesh2mesh. We show the results of different kernel sizes and the different numbers of convolutional layers. The MPJPE errors are reported. }
\label{tab:mesh2mesh-exp} 
\vspace{-0.1in}
\end{table}

\noindent\textbf{Ablation study of mesh2mesh.}
We explore different kernel sizes for the mesh2mesh module and different numbers of convolutional layers. SMPLify-X is taken as the reconstruction module and the results are reported in Table \ref{tab:mesh2mesh-exp}. 

\textbf{1) How to design the kernel size.} The size of the input mesh cuboid is $T\times N \times 3$, where $T$ means the number of meshes contained in the sequence, $N$ means the number of vertices of a single mesh, and 3 means the dimensions of vertex location. The kernel of $5 \times 3 \times 3$, $3 \times 1 \times 1$, $5 \times 1 \times 1$ and $5\times 1 \times 3$ sizes are explored as shown in Table \ref{tab:mesh2mesh-exp}. Some conclusions can be drawn from the results: 1) The kernel of $5\times 3 \times 3$ size performs worst, even worse than the SMPLify-X backbone (136.4). It is not difficult to understand because the vertices are not arranged in the actual connection state, thus such a kernel will destroy the initial mesh. 2) The kernel of $3 \times 1\times 3$ size is not as good as ours $5 \times 1 \times 3$, since a bigger stride in the temporal axis means we can utilize more adjacent meshes to extracting the temporal information. 3) The kernel of $5\times 1 \times 1$ size is also inferior to ours $5 \times 1 \times 3$, which may seem a little surprising. We attribute this improvement to the fact that we have a relatively larger receptive field.

\textbf{2) How many layers are needed for mesh2mesh.} We set the number of convolutional layers as 3, 8, and 12. Notably, since the 12-layer mesh2mesh has many more parameters to learn, we optimize it with another 600 epochs with a learning rate of $3 \times 10^{-3}$. We observe that 3-layer mesh2mesh can reach a relatively good result with an MPJPE error of 129.5, and 8-layer mesh2mesh is the most suitable choice for our task, while 12-layer mesh2mesh performs worst even we train it for 1200 epochs in total.


\section{Conclusion}

In this paper, the \emph{mesh-based action imitation} task which aims at teaching an arbitrary identity mesh to perform the same actions shown in video demonstrations is introduced. To achieve this goal, a novel \emph{M-VAI} method is proposed by us. Our framework is able to reconstruct a consistent mesh sequence and synthesize new high-quality mesh sequences by transferring the dynamics to the target identity meshes.

\section{Acknowledgement}
This work was supported by National Natural Science Foundation of China under Grant 62032006. We would like to thank Jiashun Wang and Chao Wen for their valuable help.

\bibliographystyle{ACM-Reference-Format}
\balance
\bibliography{references}

\end{document}